\documentclass{article}

\usepackage[preprint,nonatbib]{neurips_2022} % To produce the REVIEW version
\usepackage{graphicx}
\usepackage{amsmath}
\usepackage{amssymb}
\usepackage{booktabs}
\usepackage{cases}
\usepackage{multicol}
\usepackage{algorithmic}

\usepackage[utf8]{inputenc}
\usepackage[T1]{fontenc}
\usepackage{amsmath,rotating,amssymb,amsfonts}
\usepackage[english]{babel}
\usepackage{xcolor}
\usepackage[ruled,vlined]{algorithm2e}
\usepackage{graphicx}
\usepackage[titletoc]{appendix}
\usepackage{comment,tikz}
\usetikzlibrary{fit, arrows.meta}
\usepackage{enumitem}
\usepackage{xfrac}
\definecolor{green}{rgb}{0,1,0}
\definecolor{red}{rgb}{1,0,0}
\definecolor{blue}{rgb}{0,0,1}
\definecolor{yellow}{rgb}{1,.8,}
\definecolor{pink}{rgb}{1,0,1}
\definecolor{cyan}{rgb}{0,1,1}
\definecolor{black}{rgb}{0,0,0}
\definecolor{dgrey}{rgb}{.45,.26,0}
\definecolor{cgrey}{rgb}{.33,.33,.33}
\newcommand\thicklinelegend{1.2pt}
\newcommand\legendtextsize{\footnotesize}
\newcommand{\bb}{\mathbb}

\newcommand{\R}{\bb R}

\newcommand{\parent}{{\mbox{{\scriptsize parent}}}}
\newcommand{\loss}{{\mbox{{loss}}}}
\newcommand{\model}{{\mbox{{model}}}}

\usepackage[pagebackref,breaklinks,colorlinks]{hyperref}

\usepackage[capitalize]{cleveref}
\crefname{section}{Sec.}{Secs.}
\Crefname{section}{Section}{Sections}
\Crefname{table}{Table}{Tables}
\crefname{table}{Tab.}{Tabs.}

\newcounter{pbtcounter}[figure]

\begin{document}

\title{Genealogical Population-Based Training for Hyperparameter Optimization}

%\author{%
%  Anonymous author(s) \\
%   Anonymous institution(s)\\
%  \texttt{anonymous email(s)} \\
% }
\author{%
  Antoine Scardigli \\
  ENS, PSL Research University, Paris, France \\
  \texttt{antoine.scardigli@ens.psl.eu} \\
  \& \\
  Paul Fournier\\
  ENS, PSL Research University, Paris, France\\
  \texttt{paul.fournier@ens.psl.eu} \\
  \& \\
  Matteo Vilucchio \\
  ENS, PSL Research University, Paris, France \\
  \texttt{matteo.vilucchio@ens.psl.eu} \\
  \& \\
  David Naccache \\
  ENS, PSL Research University, Paris, France \\
  \texttt{david.naccache@ens.fr}
}

%\and

\maketitle 

\begin{abstract}
HyperParameter Optimization (HPO) aims at finding the best HyperParameters (HPs) of learning models, such as neural networks, in the fastest and most efficient way possible. Most recent HPO algorithms try to optimize HPs regardless of the model that obtained them, assuming that for different models, same HPs will produce very similar results. We break free from this paradigm and propose a new take on preexisting methods that we called Genealogical Population Based Training (GPBT). GPBT, via the shared histories of "genealogically"-related models, exploit the coupling of HPs and models in an efficient way. We experimentally demonstrate that our method cuts down by 2 to 3 times the computational cost required, generally allows a 1\% accuracy improvement on computer vision tasks, and reduces the variance of the results by an order of magnitude, compared to the current algorithms. Our method is search-algorithm agnostic so that the inner search routine can be any search algorithm like TPE, GP, CMA or random search.%, and has a sublinear regret if the search algorithm used has a sublinear regret.
\end{abstract}

\section{Introduction}\label{sec:Introduction}

\subsection{Related work on algorithms searching for a constant set of HPs}\label{subsec:realted-work}
During the design of learning models, the selection of many hyperparameters (HPs) within a search space often very huge is crucial for the performances obtained and for the acceleration of learning processes \cite{hinz2018speeding, albelwi2016automated, albelwi2017framework, loshchilov2016cma, talathi2015hyper}.

The main goal of Hyper-Parameter Optimization (HPO) is to develop techniques capable of finding the best HPs of a learning model with a reasonable computational cost. HPO also increases the reproducibility and reliability of results thanks to the automation of the process. HPO is challenging because it requires finding a minimum in a complex space of a computationally expensive, irregular function whose gradient is unavailable.

The first HPO method introduced was grid-search, which trivially evaluates the model's performance on series of HPs arranged in a grid in the HP space. A widespread alternative is RandomSearch \cite{bergstra2012random}, where HPs are sampled randomly in the HP space. The main shortcoming of these approaches lies in their simplicity: they do not make use of the performances of previously evaluated HPs.

Model-based algorithms, such as BO (Bayesian Optimization), aim at modelling performances of HPs on the search space. BO creates a probabilistic surrogate model on the search space and decides what HPs to evaluate next with the help of an acquisition function. A surrogate model can be, for example, GPs (Gaussian Processes) that perform well on few data points and simple hyperspaces \cite{eggensperger2013towards}, or random forests or TPE (Tree Parzen Estimator) that perform well on a more significant number of data points and complex spaces \cite{bergstra2011algorithms}.

Model-free algorithms such as Population Based (PB) methods present the advantage of being easy to parallelize as they train independently many learning models with different hyperparameters. Such algorithms are inspired by biological evolution and can use reproduction, mutation, recombination, or selection depending on models' performances. Another example, CMA-ES samples HPs configurations according to a multivariate Gaussian process, whose mean and covariance matrix are updated every iteration as a function of the best individuals of the last iteration \cite{loshchilov2016cma}.

More recent methods estimate performance on sub-datasets or on a smaller number of iterations, resulting in shorter training times. EPS \cite{provost1999efficient} predicts the learning curve's shape and uses it to abort earlier the training of a learning model with a given HPs configuration if it is predicted that this model will never perform well enough. A richer approach stemming from the meta-learning field \cite{chandrashekaran2017speeding} uses information from previously evaluated HPs configurations to adapt the learning curve predictor. It is possible not to terminate the learning model forever as it can be temporally frozen, and BO can decide at each step to either explore a new configuration or exploit (thawing) one of the frozen models with his associated HPs configuration. This approach is proposed by \cite{swersky2014freezethaw}.

Successive-halving, introduced in \cite{jamieson2016non}, evaluates many HP configurations with a fraction of the budget, then inductively keeps the best half of the models, and doubles for each one of them the training budget until only one best configuration remains. This creates an optimization up to $n /\log(n)$ times faster, with $n$ the total number of configurations.

HyperBand \cite{li2017hyperband} notices that when successive-halving trains all HPs configurations with a fraction of the budget, the fraction needs to be tuned: if it is too big, the training will be too long, and if it is too small, the selection of the best model will be wrong as performances might not be significant yet. HyperBand proposes to split the total budget into several combinations with different fraction budgets and a different number of configurations each, and then calls successive halving on all of these combinations. BOHB \cite{falkner2018bohb} proposes a variation of HyperBand, which uses Bayesian optimization instead of random search to find HP configurations. 
%There are many other approaches. Some can be problem-specific like \cite{jomaa2019hyprl}, others are more efficient ways of doing grid search \cite{lakhmiri2019hypernomad}, or rephrasing the problem in terms of Lagrangian optimization \cite{sinha2020gradientbased}.

We particularly recommend \textsc{Feurer}'s survey on the subject  \cite{Feurer2019}.

\subsection{Related work on algorithms searching for a schedule of HPs: Adaptive approaches}

Traditional HPO methods described up to now follow the suboptimal strategy of searching for a constant set of HPs to be used during the whole training.
Recent optimization algorithms propose an adaptive optimization that will discover a schedule of HPs during the training instead. By this approach, HPs of models are periodically mutated during their training. PBT (Population-Based Training \cite{jaderberg2017population}) makes the mutations randomly in the HP space. Derivative works of PBT exploit the periodical evaluations and mutations to add a time dimension to the probabilistic surrogate model on the Hyperparameter search space in order to make more insightful predictions. PB2, BOIL, and others \cite{parkerholder2021provably, nguyen2021bayesian, raj2020improving, wang2021cost} use for this purpose a Time-varying Gaussian process bandit optimization \cite{bogunovic2016time}, while \cite{angeland2020improving} uses adaptive Differential Evolution methods SHADE \cite{tanabe2013success} and LSHADE \cite{tanabe2014improving}.

It has been shown that adaptive approaches, sometimes called hyperparameter schedule search methods (see \cite{li2021automl}), are the state of the art in HPO for computer vision tasks \cite{dalibard2021faster, nguyen2021bayesian, jaderberg2017population} and more generally in deep learning and deep reinforcement learning tasks \cite{jaderberg2019human, espeholt2018impala}.
Adaptive optimization methods also reduce the bias due to noise between validation and test loss as the HP configurations are evaluated several times. 

\subsection{Problem statement}\label{sec:problem-statement}

In essence, statistical learning problems are bi-level optimization problems on $\cal H$ the space of all HPs, and $ \cal M$ the space of the models to optimize (such as, but not limited to, the weights of a neural network) that can be formulated as follows:

We define the loss $\ell : \cal M \rightarrow \R_+$ that measures how far a model is from the optimal one. 

A model $m \in \cal M$  can be optimized through the \emph{training function} $\cal T:(\mathcal M \times \mathcal H) \rightarrow \mathcal M$. In practice $\mathcal T$ represent one learning iteration, which corresponds to $T_g$ epochs ($T_g \in \mathbb R^+)$.

The \emph{search function} $\cal S :  \mathcal (\mathcal H \times \R_+ )^* \rightarrow \mathcal H$ outputs HPs given previous evaluations of HPs.\\
We define $P_m$ the set of all couples (HPs, evaluation of HPs) obtained after $m$ iteration of $\cal S$.

We show in Algorithm \ref{alg:1} and Algorithm \ref{alg:2} the pseudo-code of a non adaptive and adaptive HPO algorithm, when the budget consists of training $n$ models $t_{\max}$ times. $m_0$ is an  \textit{a priori} model (for neural networks it could be a random weight initialisation model or a pretrained model).

\begin{minipage}{.49\textwidth}
  \begin{algorithm}[H]\small
      \label{alg:1}

    \DontPrintSemicolon
    \SetAlgoLined
    \caption{Constant or non adaptive HPO}
    $P_0 \leftarrow \emptyset$ \hfill \\
    \For{ $k \in \{0,\dots, n-1\}$ }{
        $h \leftarrow  \mathcal S(P_k)$ \\
        $\model \leftarrow \mathcal T(\cdot,h)^{t_{\max}}(m_0)$\\ %T(m_0,h)^{t_{\max}}
        $\loss \leftarrow   \ell(\model)$\\
       $P_{k+1}\leftarrow P_k \cup (h, \loss)$ \\
    } 
    \vspace{1.54cm}
  \end{algorithm}%
\end{minipage}%
\begin{minipage}{.49\textwidth}
  \begin{algorithm}[H]\small
      \label{alg:2}

    \DontPrintSemicolon
    \SetAlgoLined
    \caption{Adaptive HPO}
    $P_0 \leftarrow \emptyset$ \hfill \\
    $m \leftarrow$ the  ($t_{\max} \times n$) matrix of models. \\ % \%Comment: We note $m_i^j = m[i][j]$\\
    \tcp{We note $m_i^j = m[i][j]$}
    \For{ $i \in \{0,\dots, t_{\max}-1\}$}{
    \For{ $k \in \{0,\dots, n-1\}$ }{
        $h \leftarrow  \mathcal S(P_i)$ \\
        $m_{i}^k \leftarrow \mathcal T(m_{i-1}^z,h)$ for some $z < n$\\
        $\loss \leftarrow   \ell(m_{i}^k)$\\
       $P_{i+1}\leftarrow P_i \cup (h,\loss)$ \\
    } 
    }
  \end{algorithm}
\end{minipage} \\
We observe that adaptive HPO algorithms can optimize a list of HPs $(h_i)_{1\le i\le t_{\max}} \in \mathcal H^{t_{\max}}$ and that non-adaptive optimization is a particular case of adaptive-optimization with constant $(h_i)_{i\le t_{\max}}$.\footnote{If some HPs need to remain constant, such as architectural HPs, they can be fixed constant for generation steps $i > 0$. \ \ \ Also, in the adaptive case, HPs and their evaluations stored in $P_k$ need to be put in the temporal context they were obtained. For this reason, we often extend the hyperparameter search space from $\mathcal{H}$ to $\mathcal{H} \times t_{\max}$ and store $((h,k),x)$ in $P_k$ rather than $(h,x)$, where $h$ is a HP set, $k$ the generation step at which it was evaluated, and $x$ the result of the evaluation. This way evaluations from very old generation step can have less impact, and temporal patterns can be learned. \\}

Finally, we introduce the function $f_m : \mathcal{H} \rightarrow \R_+ $ s.t. $ f_m(h) =  \ell(\mathcal{T}(m,h))$ in the adaptive case, and $\ell(\mathcal T(\cdot,h)^{t_{\max}}(m))$ in the non-adaptive case. This is the function that $\cal S$ tries to approximate and minimize using previous HPs evaluations contained in $ P$.

The main contribution of this paper is to notice and find a solution for the following. We can see that in the non adaptive approach, $P_k$ contains $k$ observations of the same function $f_{m_0}$ which are used by $\mathcal{S}$ to make the next predictions. However, in adaptive approaches, at generation step $i \le t_{\max}-1$, all models $m_{i}^k$, $k < n$ are different (in the deep learning field they would have different weights) because they have been trained from HP optimization step $0$ to $i-1$ with different HPs. Since all models $m_{i}^k$ are different, they each have a different function $f_{m_{i}^k}$. This is a problem because it implies that $\mathcal{S}$ uses observations from these different functions $f_{m_{i}^k}$ to minimize a different function $f_{m_{i}^{k'}}$. There are no hence no guarantees for good performances.

Indeed, all recent adaptive optimization approaches like PB2, BOIL, and others \cite{parkerholder2021provably, nguyen2021bayesian, raj2020improving, wang2021cost, angeland2020improving} make this implicit assumption that different models react similarly to hyperparameters in their training. More formally, they assume that all functions $f_{m_{i}^k}$, $k < n$ are equal up to a (negligible) noisy term for a given generation step $i$. Thanks to this assumption, all HPs evaluations from all previous models evaluated are indistinctly taken into account: new HPs are chosen using the whole set of observations $P_i$. We from now on call this assumption the \textit{isomodel} HPs assumption. 

We will present our approach in the next section which breaks free from the isomodel HPs assumption, and we will show in the experimental section that breaking free from this assumption causes serious quality and performance improvements.

\section{Genealogical Population-Based Training}\label{sec:contribution}

We will present the mechanisms of GPBT, our proposed approach to challenge the isomodel HPs assumption, \textit{i.e.} the assumption that all functions $f_{m_{i}^j}$, $k < n$ are equal for a given generation step $i$, or equivalently, that different models react similarly to the same HPs. We will on the contrary consider the case where different models can react differently to HPs hence all have different functions $f_{m_{i}^j}$.

Our approach is a single-run adaptive optimization approach like PBT and PB2 (see Algorithm \ref{alg:2}). Indeed the general framework is the following: we divide the training of models into several ($t_{\max}$) generation steps. Each generation step starts by dropping the worst performing models from the last generation step and replicating the best ones such that the number of models currently training is constant. Then, all models are given new HPs chosen by $\cal S$. The generation step ends with all models being trained by $\cal T$, \textit{i.e.} being trained for $T_g$ epochs with the current HPs.

Current approaches making the isomodel HP assumption use $n + n\cdot(i-1)$ HPs evaluations to approximate a single function $f_{m_i}$ per generation step. The HPs evaluations come from the current generation step and from previous generation steps contained in $P_i$. \\% even if all HPs evaluation used are obtained from evaluating different models because the isomodel HPs assumption allows it. 
An implication of removing the isomodel HPs assumption is that the number of previous HPs evaluations that are relevant for approximating $f_{m_{i}^j}$ is drastically reduced. % On the contrary, the difficulty that arises when one challenges that isomodel HPs assumption is that the number of previous HPs evaluations that are relevant for approximating $f_i^j$ is drastically reduced. 
Indeed it implies to aim at approximating the $n$ different functions per generation step $f_{m_{i}^j}$, $j < n$ using only $n$ HPs evaluations that are each relevant for only one function. This challenge is solved in our approach by reducing by a factor $\sqrt{n\cdot c}$ the number of approximated functions $f_{m_i^j}$ for a given generation step $i$, which will allow every function to be approximated using $\sqrt{n \cdot c}$ times more HPs evaluations: In our method, we select the $\sqrt{n/c}$ best performing models after every generation step (called \textit{parents}). Each parent replicates itself $\sqrt{n \cdot c}$ times to produce the next generation step (the $\sqrt{n \cdot c}$ instances are called \textit{children} of the original parent). Here $c$ can be seen as the ratio between the number of children per parent and the number of parents.\footnote{$c$ is constant during the algorithm, but one can choose its value as a function of $n$ such that the number of parents is not necessarily proportional to the square root of $n$. However, we will from now on assume that $\sqrt{n/c} = \Theta(\sqrt n)$  for the sake of simplicity.} We show in the Appendix that varying the value of $c$ has negligible impact on the performance and suggest a way to set it automatically and dynamically. Furthermore, as we reject the isomodel HPs assumption, $\cal S$ should not be allowed to use $P_{i-1}$ (HPs evaluations from previous generation steps) at generation step $i$ to make its predictions, because all functions $f_{m_p^j}$, $j < n, p < i$ are expected to be different to the functions $f_{m_i^j}$. The implementation in which we only consider HPs evaluations from other children of the parent of a model will be referred to as \mbox{\textit{GPBT no-time}}.

\SetKwComment{Comment}{$\triangleright$\ }{}
\SetKwInput{KwInput}{Input}         
\SetKwInput{KwOutput}{Output}
\begin{algorithm}[!htbp]\small
    \DontPrintSemicolon
    \SetAlgoLined
    \caption{GPBT Algorithm}
    \label{alg:GPBT}
    \KwInput{\textbf{$n$} the number of children per generation step,     $\cal T$: trainer,     $\cal S$: HP search algorithm,     $t_{\max}$: maximum number of generation steps,     $T_g$: number of training iterations between two generation steps,     $c$: number of children per parent over number parents ratio,     $m_0$: initial model.}
    %\KwOutput{Best performing model and associated HP schedule $(h_0, h_1, ..., h_{t_{\max}-1})$}
    % \KwData{$\mathcal T_1$, $\mathcal T_2$}
  
    %\tcp{First generation}
    $P_0 \leftarrow \emptyset$ \quad\Comment*[h]{S's search history initialized} \\
    \For{$k \in \{0,\dots n-1 \}$ }{
         $h_k \leftarrow \mathcal S(P_0)$\quad\Comment*[h]{Search once for HPs with the initial search history}\\
         $m_k \leftarrow \mathcal T (h_k,m_0)$\quad\Comment*[h]{kth child training}\\
         $\loss_k  \leftarrow \ell (m_k)$\quad\Comment*[h]{kth child evaluation}\\
         $P_0  \leftarrow P_0 \cup (m_k, h_{k}, \loss_{k})$\quad\Comment*[h]{Update search history}\\
    }

    Select in terms of $loss_k$ the $\sqrt{n/c }$ best children, that becomes the new parents. Allocate a copy $P_0^k$ of $P_0$ to each parent $k$\\
%    $h_0 \leftarrow h_{\argmin_k \{\smallloss_k\}_{k<n}}$ \\
    
    \For{ $i \in \{1,\dots, t_{\max-1}\}$ }{
        \For(\quad\Comment*[h]{Iterate over all the parents}){ $z\in \left\{0,\dots,\sqrt {n/c}\right\}$}{
            $m_i^{\parent} \leftarrow$ model from the $z$-th parent \\
            $P_i^{\parent} \leftarrow$ ancestry line from the $z$-th parent (or $\emptyset$ if GPBT no-time)\\
            \For(\quad\Comment*[h]{Iterate over the children}){$d\in \left\{0,\dots, \sqrt{n \cdot c}\right\}$}{
                    $j \leftarrow \sqrt{n \cdot c} \cdot z + d$ \quad\Comment*[h]{unique id for every children} \\
                    $h_{i+1}^j \leftarrow \mathcal S(P_i^{\parent})$ \quad\Comment*[h]{Search HPs set given search ancestry line}\\
                    $m_{i+1}^j \leftarrow \mathcal T (h_{i+1}^j, m_i^{\parent})$ \quad\Comment*[h]{kth child is trained}\\
                    $\loss_k \leftarrow \ell (m_{i+1}^j)$ \quad\Comment*[h]{jth child is evaluated}\\
                    $P_i^{\parent} \leftarrow P_i^{\parent} \cup (m_{i+1}^j,h_{i+1}^j,\loss_j)$ \quad\Comment*[h]{Ancestry line updated}
            }
        }
    Select in terms of $\loss_j$ the $\sqrt{n/c}$ best performing models among the children and the parents, that become the new parents. Allocate to every new parent a copy of the ancestry line $P_i^{\parent}$ of their respective parent\\ 
    }

  \textbf{Output}: the best $m_{t_{\max}}^j$ and its HP schedule $(h^j_i)_{i<t_{\max}}$
\end{algorithm}

\begin{figure}[!htbp]
    \centering
    \includegraphics[width=.5\linewidth]{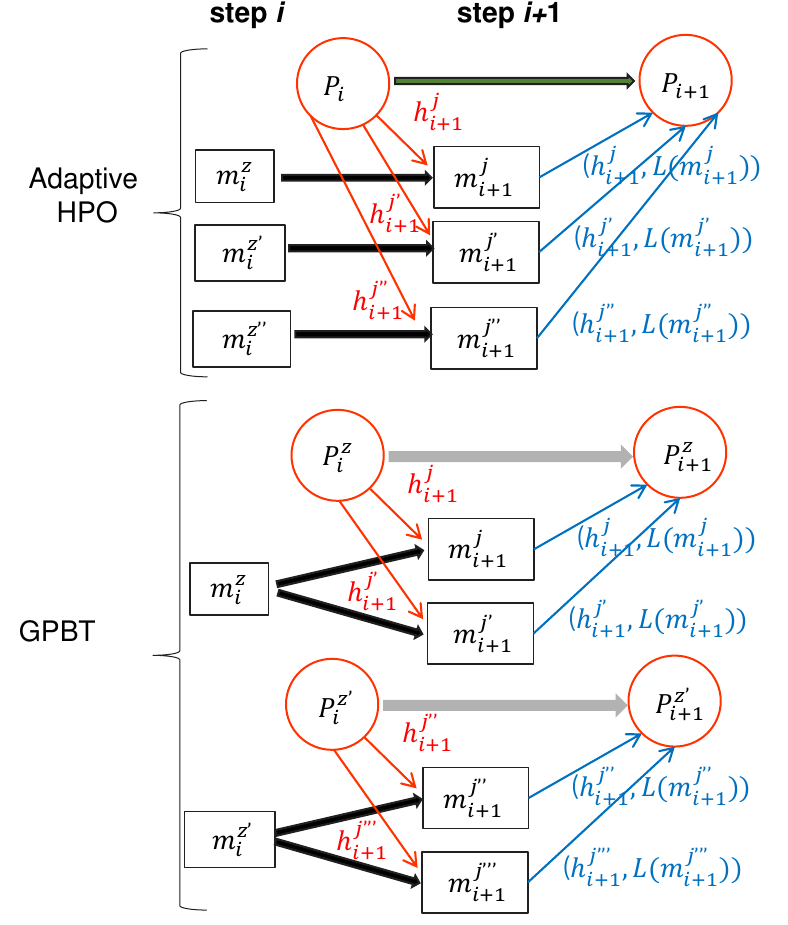}
        \caption{Comparison of adaptive HPO approaches with ours. We show the training of 3 models in for the adaptive HPO approach, and 2 parent models training each 2 children models for our approach. Black arrows represent the training of a model. In the adaptive case, trainings happen with HPs predicted using the same $\mathcal{P}_i$ (see red arrows and Algorithm 2) and evaluations of the HPs are gathered back to $\mathcal{P}_{i+1}$ in a model agnostic way (see blue arrows and Algorithm 2) therefore not taking into consideration that  different models may train and evaluate very differently HPs. In our approach, HPs predictions (red arrows) and HPs evaluations (blue arrows) are done with $\mathcal{P}_{i+1}^z$ that is distinct for every parent model $z$, therefore predictions are robust to the case where different models react differently to HPs. The grey arrow symbolizes that $\mathcal{P}_{i+1}^z$ originates from $\mathcal{P}_{i}^z$ in the GPBT-time method. In GPBT no-time, $\mathcal{P}_{i+1}$ only enriches thanks to the HPs evaluations from current generations step; hence there would not be the grey arrow. $L$ refers to $\ell$ from the problem statement section. }
    \label{fig:figure1}
\end{figure}

A compromise between the approach of GPBT no-time and making the isomodel assumption can be to assume that models with a similar history have similar functions $f$. In this case we can allow $\cal S$ to use the subset of $P_{i-1}$ that contains all HPs evaluations from models genealogically related to a model. One can choose to exploit the generation of all other children of the father or all the children of the grandfather and so on. This will be referred to as \mbox{\textit{GPBT time}}. GPBT time approximates $\Theta(\sqrt{n})$ functions per generation step, each one using $\Theta(\sqrt{n})$ HPs evaluations plus $\Theta(\sqrt{n}\cdot(i-1))$ HPs evaluations contained in $ P_{i-1}$ because we include $\Theta(\sqrt n)$ HPs evaluations per previous generation steps. This can be implemented by creating an ancestry line ${P}_i^q$, $q\le  \sqrt{n/c}$ for all $\sqrt{n/c}$ parents instead of having a single set $ P_{i-1}$ of previous evaluations. $ P_i^q$ gets updated by appending all HP evaluations from the children of the parent that possess $\ P_i^q$. All the children of this parent  that are selected for becoming parents at next generation step then get a copy of the updated ${P}_i^q$ that becomes $P_{i+1}^{q'}$, $q'\le \sqrt{n/c}$ with $q'$ the index(es) of such parent(s). See Algorithm \ref{alg:GPBT}, Table~\ref{tab:comparison} and Figure \ref{fig:figure1}. Therefore our solution breaks free from the isomodel HPs assumption because HP searches are performed independently for different models. 
\begin{table}[]
\centering
\renewcommand{\arraystretch}{2}
\begin{tabular*}{\columnwidth}{p{.5\columnwidth} p{.14\columnwidth} p{.14\columnwidth} p{.14\columnwidth}}
\hline \hline
 & Adative  & GPBT  & GPBT  \\ [-0.4cm]
 & HPO & no-time & time \\
    \hline
\#\ Approximated functions $f_{m}$ per generation steps & 1	& $\Theta(\sqrt{n})$	&$\Theta(\sqrt{n})$\\ [-0.3cm]
\#\ HPs evaluations from current generation step used per approximated function&$\Theta(n)$	&$\Theta(\sqrt{n})$	&$\Theta(\sqrt{n})$\\ [-0.3cm]
\#\ HPs evaluations of identical models from current generation step used per approximated function&$\Theta(1)$	&$\Theta(\sqrt{n})$	&$\Theta(\sqrt{n})$\\ [-0.3cm]
\#\ Previous HPs evaluations used from $\mathcal P_{i}$ per approximated function&$\Theta(n\cdot i) $ & $\Theta(1)$ & $\Theta(\sqrt{n}\cdot i)$\\[-0.3cm]
\#\ Previous HPs evaluations of genealogically-related models used from $\mathcal P_{i}$ per approximated function&$\Theta(i) $& $\Theta(1)$ & $\Theta(\sqrt{n}\cdot i)$\\
\hline \hline
\end{tabular*}

\caption{Comparison of how HPs evaluations are used to do new HPs predictions for Adaptive HPO algorithms and ours at generation step $i+1$.}
\label{tab:comparison}
\end{table} \\
Intuitively, our approach is strong in exploitation because only the $\sqrt{n/c}$ best performing models are kept after every generation step, and strong in exploration because it maintains the approximation of $\sqrt{c \cdot n}$ functions  $f_{m_i^j}$ instead of only $1$. Its main weakness is that the small number of HPs evaluations might not be enough to approximate each $f_{m_i^j}$ function. In order to make our method more robust, we make two complementary improvements:
\begin{enumerate}
    \item We assign to the first children of an offspring the HPs of his father. Since the father was well performing with those HPs, we can hope that the child will perform at least as well, while other children can do the riskier exploration of hyperparameters. Also in case the child performs well with the HPs of its parent, it will guide $\cal S$ to make better predictions.
    \item Instead of doing the selection of the best $\sqrt{n / c}$ performing models among the $n$ children in order to get the parents for the next generation step, we do the selection among the $n$ children and their $\sqrt{n / c}$ parents. This guarantees firstly that performances will never drop, but even more importantly that if a very well performing parent is successively selected $b$ generation steps, this parent will have had $b\cdot \sqrt{n\cdot c}$ children, and hence will be able to use a lot of HPs evaluations for new HPs predictions.
    \end{enumerate}

We implemented GPBT in a modular way  such that any non-adaptive HPO algorithms such as RandomSearch, BO-TPE, BO-GP or CMA-ES can be used as the internal search algorithm $\cal S$.
As explored in the experimental section, GPBT's only critical external parameter is the choice of the search algorithm $\cal S$. The influence of some parameters such as $c$ is negligible and can be dynamically optimized (see Appendix). The choice of $T_g$ mostly depends on the number of iterations needed by $\cal T$ to get reliable results, which is very important for doing a reliable selection of the best parents. The parameters $t_{\max}$ and $n$ reflect the computational budget that one wants to allocate. % Experimentally, this has never been an issue even when setting small $T_g$.
Our algorithm can run in parallel up to $\sqrt{n/c}$ instances of the search algorithm $\cal S$ (the number of parents). If $\cal S$ is itself parallelizable, all $n$ models can be trained in parallel, allowing a single run optimization with execution time $\mathcal{O}(t_{\max} \cdot t(\mathcal T)) = \mathcal{O}(t_{\max} \cdot  T_g \cdot t(\text{epoch}))$, where $t(\mathcal T)$ represents the time complexity of $\mathcal T$, which is $T_g \cdot t(\text{epoch})$, and $t(\text{epoch})$ is the time complexity of one learning iteration (epoch).

\textbf{Early Stopping:}
In the Appendix, we describe three complementary levels of early stopping. We implemented the finest-grained one of the three which consists of early-stopping the training of a child if it is not likely that it will perform better than other children. More precisely, every time a child is trained through $\cal T$, it is early-evaluated after one epoch, and stopped if its performances are worse than the median of other children already early-evaluated \footnote{This does not affect parallel/distributed training as a simple lock on the early evaluations can be used.}.\\
In the Appendix, we prove that this early stopping strategy causes an almost 2-fold acceleration lower bound. In the experimental section, we empirically observe a 2-3 fold acceleration for HPs tuning of computer vision tasks at a small performance cost.  

\textbf{Appendix:}
The Appendix contains insightful additions such as:
\begin{itemize}
    \item A comparison study of GPBT time vs GPBT no-time
    \item A toy HPs search space visualisation comparing GPBT with PBT
    \item A study of the relative effect of the genealogical and search function contribution

\end{itemize}

\section{Experimental results}\label{sec:experimental-results}
We evaluate our approach using several diverse search algorithms $\cal S$:  GPBT-CMA, GPBT time-GP (using time-varying Gaussian process bandit \cite{bogunovic2016time, srinivas2012information}), GPBT-GP, GPBT time-TPE (Tree Parzen Estimators) and GPBT-TPE.\\
Compared baselines include two widely-used non-adaptive approaches: Hyperopt, relying on TPE, and BOHB, relying on GP and HyperBand, and the two most used adaptive approaches: PBT and PB2. We do not use CMA as a baseline although we used it as a search algorithm because it turned out to perform badly in our experiments due to the small population size. We progressively reduce the number of variations of GPBT and of baselines to only keep the ones performing well as the computational costs increase. In order to make our experiments as diverse as possible, we test different ranges of search spaces and different numbers of HPs for each experiment. More experiment details are available in the Appendix. All compared algorithms are given the same constraints in terms of maximal number of iterations and maximal number of models trained per iteration. We choose as metric the \textit{mean best seen} validation and corresponding test accuracy/error/FID as a function of the wall clock time (including training time, evaluation time, HPs predictions, model transfers...) because it is one of the most used metrics used for comparisons in HPO \cite{falkner2018bohb} \cite{li2017hyperband} \cite{bergstra2011algorithms}. %reference?

\subsection{Toy Boston experiment and suboptimality of constant HPO }
We use a simple multi-layer perceptron on normalized Boston \cite{Harrison1978} with AdaBelief \cite{zhuang2020adabelief} optimizer and aim at optimizing $6$ HPs with $n=72$ learning models, $t_{\max}=5$ generation steps, $c=1$, $T_g=1$ (one epoch per generation step). We do not use the speed-up implementation for GPBT.

\begin{figure}[!htbp]
    \centering
    \includegraphics[width=1\linewidth,height=.25\linewidth]{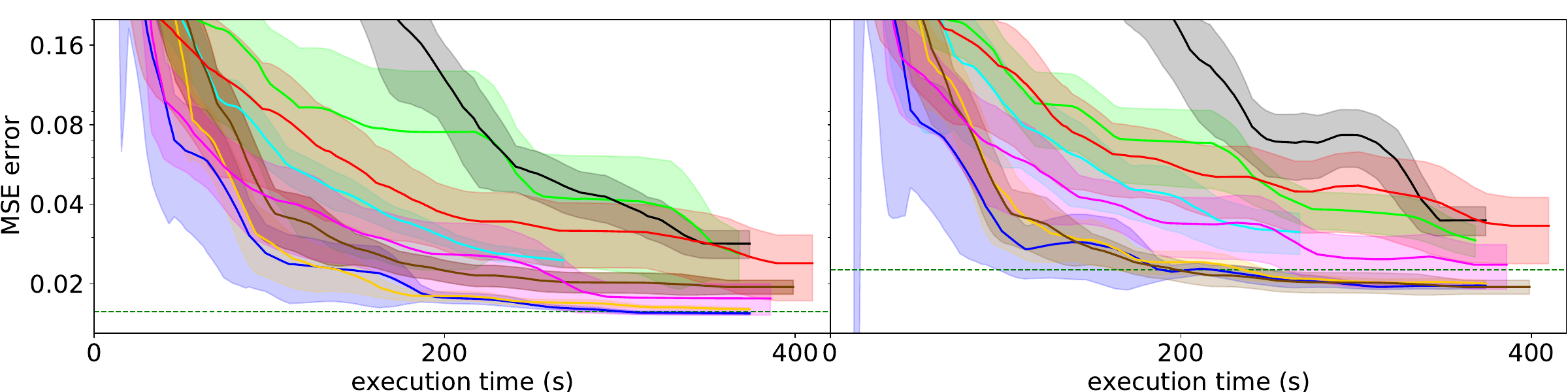}
    
    \begin{tabular}{|cccc|}\hline
        \textcolor{red}{\legendtextsize PB2} & 
        \textcolor{green}{\legendtextsize HyperOpt} & 
        \textcolor{pink}{\legendtextsize PBT} &
                \textcolor{cyan}{\legendtextsize BOHB} \\ [-0.3cm]
        \textcolor{red}{\rule{1cm}{\thicklinelegend}} & 
        \textcolor{green}{\rule{1cm}{\thicklinelegend}} & 
        \textcolor{pink}{\rule{1cm}{\thicklinelegend}} &
        \textcolor{cyan}{\rule{1cm}{\thicklinelegend}} \\
        \textcolor{blue}{\legendtextsize GPBT time-TPE} & 
        \textcolor{yellow}{\legendtextsize GPBT-TPE} &
        \textcolor{black}{\legendtextsize GPBT-CMA} & 
        \textcolor{dgrey}{\legendtextsize GPBT time-GP}  \\ [-0.3cm]
        \textcolor{blue}{\rule{1cm}{\thicklinelegend}} & 
        \textcolor{yellow}{\rule{1cm}{\thicklinelegend}} &
        \textcolor{black}{\rule{1cm}{\thicklinelegend}} & 
        \textcolor{dgrey}{\rule{1cm}{\thicklinelegend}}  \\ \hline
    \end{tabular}
    \caption{Comparison of val (leftmost) and test (rightmost) best seen MSE error on Boston with 10 experiments. Lines represent means, and shaded-areas the STDs. We used logarithmic scale on the y-axis. The green dashed line gives a bound on the optimal performance obtainable by non adaptive algorithms, here simulated by 7200 RandomSearch runs.}
    \label{fig:toy-boston}
\end{figure}

As we can observe in Figure~\ref{fig:toy-boston}, the choice of the search algorithm creates important variations in performance. Still, GPBT time-TPE, GPBT-TPE, GPBT time-GP dominate in terms of performance, and variance compared to other algorithms.

%The fact that not all algorithms finish at the same time despite an equal resource allocation and equal iteration constraints is due to the fact that all algorithms have different weight transferring policies, or take longer to make HPs predictions. We can observe that all adaptive approaches are slightly slower than other approaches because they do a lot of weights transfer. As we can see here and as sketched in the Appendix D, our approach is slightly faster than other adaptive approaches because weight transfers are made more efficiently.

The dashed green line represents the best validation and corresponding test error reached at the best iteration of the best model when training 7200 models with RandomSearch. Assuming that this random search extensively covered the HP search space, this gives an insight on the sub-optimality of non-adaptive approaches, because our adaptive approach outperforms the extensive random search despite training with only 72 models.
%This might also explain the minimal variance that our algorithm demonstrates: it might not only be an indicator of robustness, but it could also be a sign of optimality: our algorithm would find a near-optimal solution at every experiment explaining the small variance.

\subsection{MNIST classification}
We use LeNet \cite{LeCun1989} network on non-preprocessed MNIST \cite{LeCun1998} with Adam \cite{kingma2017adam} optimizer and aim at optimizing 5 HPs with $n=25$, $t_{\max}=10$, $c=1$, $T_g=1$. We do not use the speed-up implementation for GPBT.

\begin{figure}[!htbp]
    \centering
    \includegraphics[width=1\linewidth]{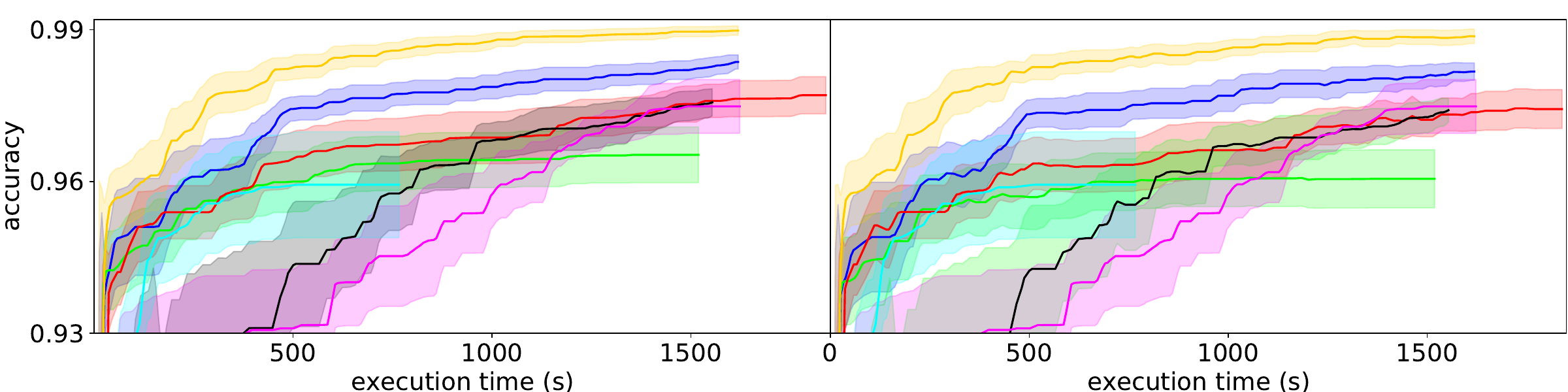}
    
    \begin{tabular}{|cccc|}\hline
        \textcolor{red}{\legendtextsize PB2} & 
        \textcolor{green}{\legendtextsize HyperOpt} & 
        \textcolor{pink}{\legendtextsize PBT} &
        \textcolor{cyan}{\legendtextsize BOHB} \\ [-0.3cm]
        \textcolor{red}{\rule{1cm}{\thicklinelegend}} & 
        \textcolor{green}{\rule{1cm}{\thicklinelegend}} & 
        \textcolor{pink}{\rule{1cm}{\thicklinelegend}} &
        \textcolor{cyan}{\rule{1cm}{\thicklinelegend}} \\
        \textcolor{blue}{\legendtextsize GPBT time-TPE} & 
        \textcolor{yellow}{\legendtextsize GPBT-TPE} &
        \textcolor{black}{\legendtextsize GPBT-CMA} & \\ [-0.3cm]
        \textcolor{blue}{\rule{1cm}{\thicklinelegend}} & 
        \textcolor{yellow}{\rule{1cm}{\thicklinelegend}} &
        \textcolor{black}{\rule{1cm}{\thicklinelegend}} &  \\ \hline
    \end{tabular}
    \caption{Comparison of validation (leftmost) and test (rightmost) best seen accuracy on MNIST with 10 experiments. Lines represent means, and shaded-areas the STDs.}
    \label{fig:toy-mnist}
\end{figure}

As we can observe in Figure \ref{fig:toy-mnist}, GPBT-TPE and GPBT time-TPE have the best results and the smaller variance for no time overhead. GPBT-CMA does not work well probably because CMA-ES is powerful only for experiments with many workers.

\subsection{CIFAR-10 and IMAGENET classification}
\begin{figure}[!htbp]
    \centering
    \includegraphics[width=1\linewidth, height =.23\linewidth ]{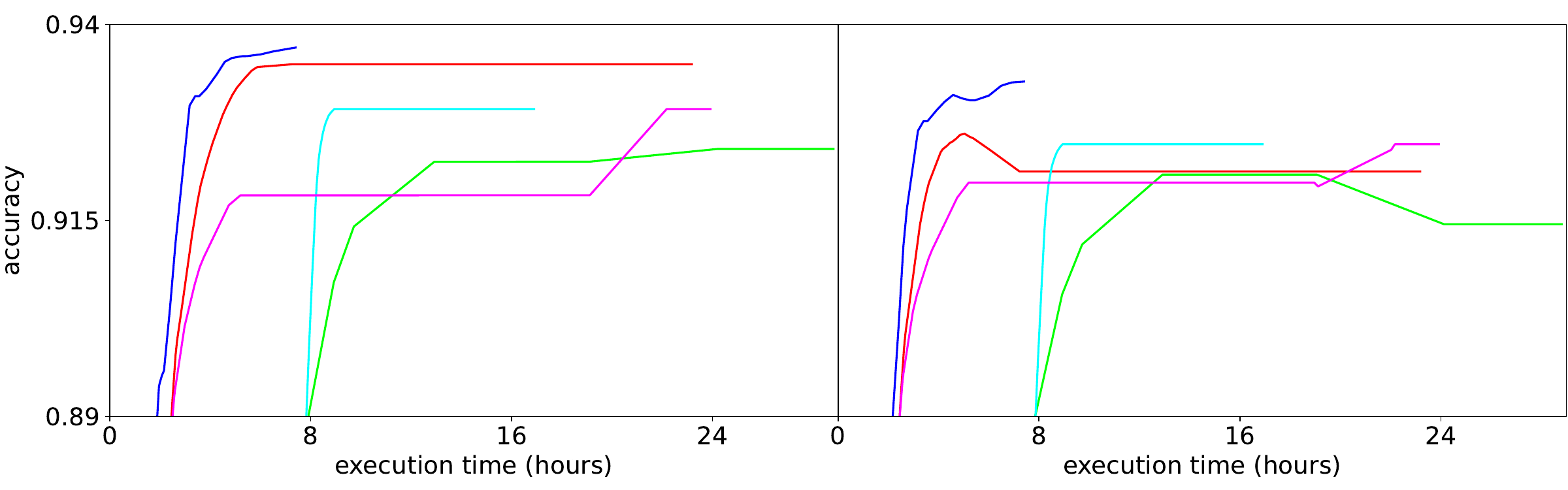}
    \begin{tabular}{|ccccc|}\hline
        \textcolor{red}{\legendtextsize PB2} & 
        \textcolor{pink}{\legendtextsize PBT} & 
        \textcolor{cyan}{\legendtextsize BOHB} & 
        \textcolor{green}{\legendtextsize HyperOpt} &
        \textcolor{blue}{\legendtextsize GPBT TPE} \\ [-0.3cm]
        \textcolor{red}{\rule{1cm}{\thicklinelegend}} & 
        \textcolor{pink}{\rule{1cm}{\thicklinelegend}} & 
        \textcolor{cyan}{\rule{1cm}{\thicklinelegend}} & 
        \textcolor{green}{\rule{1cm}{\thicklinelegend}} &
        \textcolor{blue}{\rule{1cm}{\thicklinelegend}} \\ \hline
    \end{tabular}
    % \tikz\draw[line]
    \caption{Comparison of val (leftmost) and test (rightmost) best seen accuracy for CIFAR-10.}
    \label{fig:cifar-10}
\end{figure}

We use SimpleNet \cite{hasanpour2016lets} on normalized CIFAR-10 \cite{Krizhevsky09learningmultiple} with optimizer Ada-Delta \cite{DBLP:journals/corr/abs-1212-5701} and aim at optimizing $5$ HPs with $n=36$, $t_{\max}=10$, $c=1$, $T_g=5$. The results are shown in Figure \ref{fig:cifar-10}.

\begin{figure}[h!]
  \centering
  \includegraphics[width=1\linewidth,height = .2\linewidth]{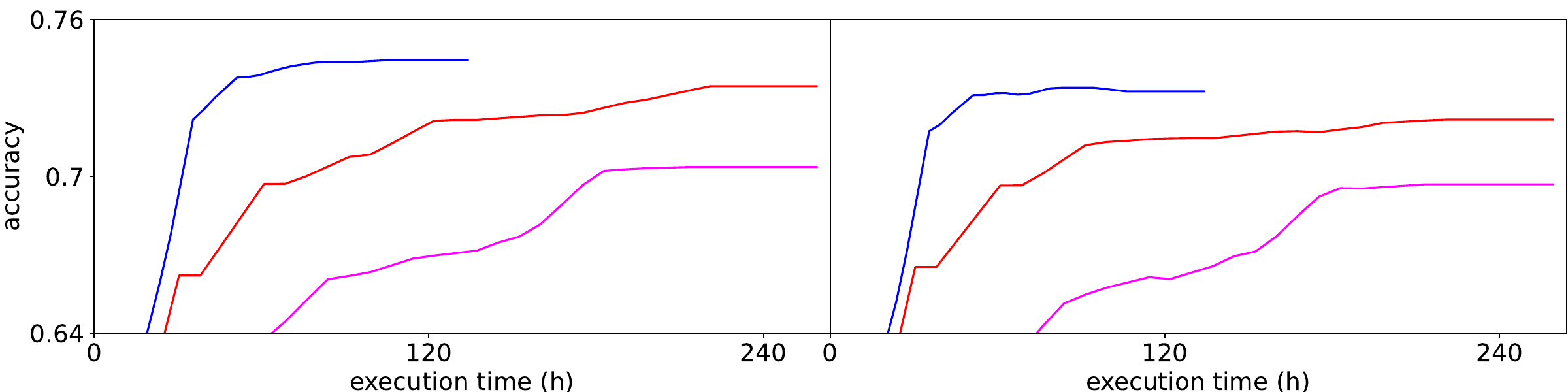}
    \begin{tabular}{|ccc|}\hline
        \textcolor{red}{\legendtextsize PB2} & 
        \textcolor{pink}{\legendtextsize PBT} &
        \textcolor{blue}{\legendtextsize GPBT time-TPE} \\ [-0.3cm]
        \textcolor{red}{\rule{1cm}{\thicklinelegend}} & 
        \textcolor{pink}{\rule{1cm}{\thicklinelegend}} &
        \textcolor{blue}{\rule{1cm}{\thicklinelegend}} \\ \hline
    \end{tabular}
   \caption{Comparison of val (leftmost) and test (rightmost) best-seen accuracy using ResNet50 on ImageNet.}
   \label{fig:first}
\end{figure}

We present in Figure \ref{fig:first} an experiment training ResNet50 \cite{he2016deep} on ImageNet-1k \cite{deng2009imagenet}.
 We used as parameters $n$ = 4, $c$ = 4, $t_{max}$ = 80, $T_g$ = 5, and optimized 1 HP. We use GPBT time-TPE for this experiments with few models ($n$=4) as we showed in the Appendix that we expect GPBT time-TPE to outperform GPBT TPE in this case.

GPBT uses in those two experiments the finest-grained level of speed up described in Section 2 and in the Appendix. The empirical acceleration is higher than the theoretical 2-fold acceleration lower-bound as we can see that our algorithm terminates between two to three times sooner than other algorithms. Despite this time-acceleration, our approach still outperforms all other approaches performance-wise. We also observe that our approach is less subject to overfitting than PB2. 

\subsection{DC-GAN on CIFAR-10 and STL-10}
\begin{figure}[!htbp]
    \centering
    \includegraphics[width=.49\linewidth,height = .23\linewidth]{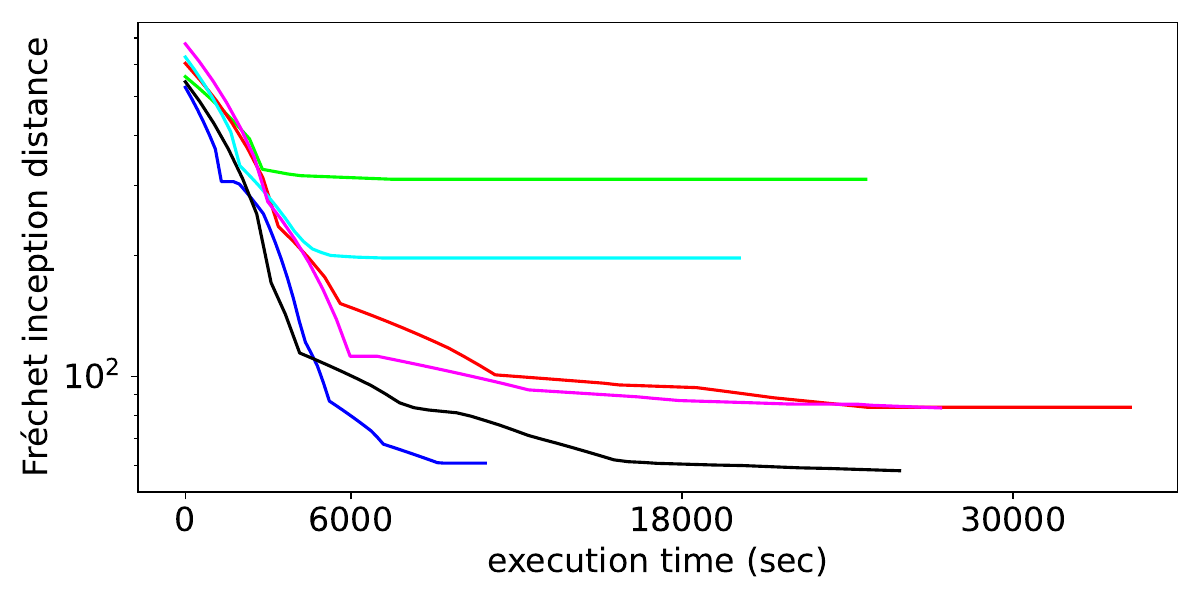}
    \includegraphics[width=.49\linewidth,height= .23\linewidth]{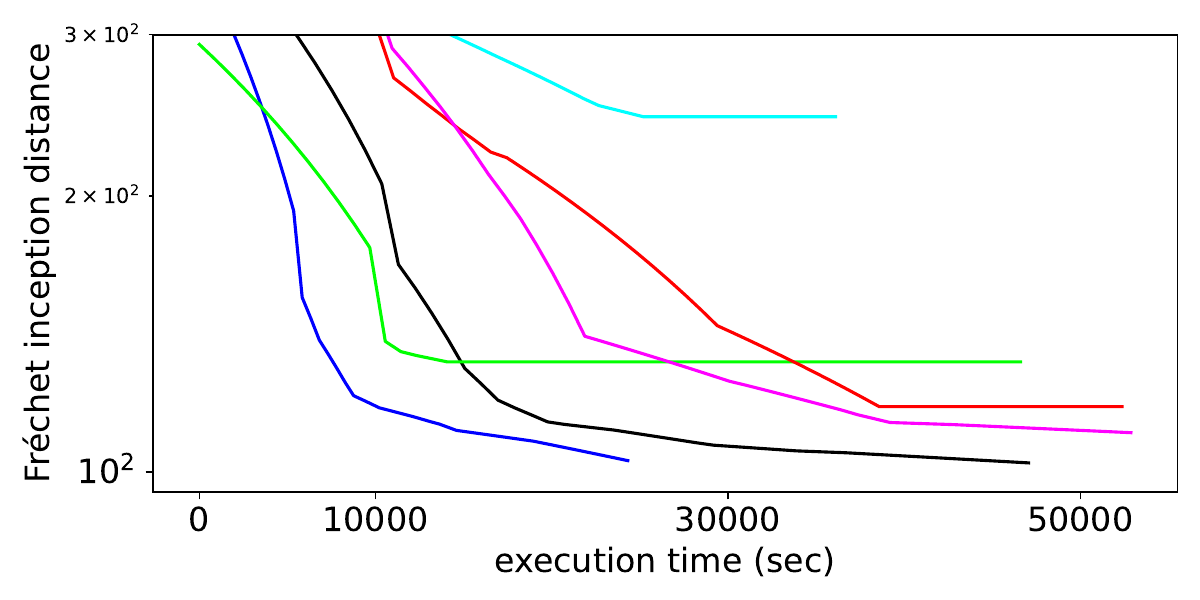}

    \begin{tabular}{|cccccc|}\hline
        \textcolor{red}{\legendtextsize PB2} & 
        \textcolor{pink}{\legendtextsize PBT} & 
        \textcolor{cyan}{\legendtextsize BOHB} & %[-0.3cm]
        \textcolor{green}{\legendtextsize HyperOpt} &
        \textcolor{blue}{\legendtextsize GPBT TPE} &
        \textcolor{black}{\legendtextsize GPBT TPE w/o speedup} \\ [-0.3cm]
                \textcolor{red}{\rule{1cm}{\thicklinelegend}} &  
        \textcolor{pink}{\rule{1cm}{\thicklinelegend}} & 
        \textcolor{cyan}{\rule{1cm}{\thicklinelegend}} &
        \textcolor{green}{\rule{1cm}{\thicklinelegend}} &
        \textcolor{blue}{\rule{1cm}{\thicklinelegend}} &
     \textcolor{black}{\rule{1cm}{\thicklinelegend}} \\ \hline
    \end{tabular}
    % \tikz\draw[line]
    \caption{Comparison of best seen Inception distance using a DC-GAN on CIFAR-10 (left) and STL-10 (right). We used logarithmic scale on the y-axis.}
    \label{fig:cifargan-10}
\end{figure}

We use an inception-DCGAN \cite{goodfellow2020generative} on normalized CIFAR-10 \cite{Krizhevsky09learningmultiple} and STL-10 \cite{coates2011analysis} with optimizer AdaBelief \cite{DBLP:journals/corr/abs-1212-5701} and aim at optimizing $4$ HPs with $n=9$, $t_{\max}=5$, $c=1$, $T_g=5$.
For these two experiments, we compare our approach with and without the finest grained-level of speed-up. The results are shown in Figure \ref{fig:cifargan-10}.

The experiments demonstrate that our approach can outperform other baselines by an important margin on non classification tasks as well, confirms the trend from the Figure \ref{fig:cifar-10} that the empirical acceleration of our approach is at least 2-fold, and shows that using our accelerated approach only causes a negligible loss in the performances compared to the non accelerated one.

\section{Conclusion}
We introduced, to our knowledge, the first adaptive algorithm that searches HPs independently for models with different weights. Our approach is also modular and independent as it does not require a specific search algorithm but can be a wrapper for any search algorithms such as TPE, GP, CMA, Random Search, or other algorithms that likely yield even better results.

Our experimental results showed that breaking free from the isomodel HPs assumption improves performances: HPs evaluations should not be used indiscriminately of the models that generated them, it is instead important to limit the access of HPs evaluations to make new HPs predictions. \\We also observed that GPBT have three advantages compared to the state of the art: It improves significantly the performances (at least 1\% in all our image classification experiments); it is at least two times faster; and it has one order of magnitude less variance in performance. 

Using GPBT will make HPs tuning less computationally expensive in the machine learning community, and probably for many other iteration-based optimization tasks. To facilitate the use of our method, we publish our implementation \href{https://anonymous.4open.science/r/GPBT-FFAD/README.md}{here} under MIT licence. We will soon propose an implementation compatible with the widely used Ray-Tune library \cite{liaw2018tune}.

%%%%%%%%% REFERENCES
\bibliography{opt}
\bibliographystyle{plain}

\end{document}